\documentclass[11pt]{article}

\usepackage[final]{acl}

\usepackage{times}
\usepackage{latexsym}

\usepackage[T1]{fontenc}
\usepackage[utf8]{inputenc}

\usepackage{microtype}

\usepackage{inconsolata}

\usepackage{graphicx}

\usepackage[normalem]{ulem}
\usepackage{amsmath}
\usepackage{amssymb}
\usepackage[table]{xcolor}
\usepackage{amsthm}
\usepackage[most]{tcolorbox}
\usepackage{dblfloatfix}

\usepackage{enumitem}
\usepackage{needspace}
\makeatletter
\newtcolorbox[auto counter]{llmrpbox}[1]{
  width=0.98\linewidth, breakable, enhanced, colback=white, colframe=black,
  boxrule=0.5pt, arc=2pt, left=5pt, right=5pt, top=4pt, bottom=4pt,
  fonttitle=\bfseries\footnotesize, fontupper=\footnotesize,
  before upper={\raggedright\setlength{\parindent}{0pt}\def\@currentlabel{\thetcbcounter}},
  title={Box~\thetcbcounter: #1}}
\makeatother
\tcbset{llmrbox/.style={
  width=0.98\linewidth, breakable, enhanced, colback=white, colframe=black,
  boxrule=0.5pt, arc=2pt, left=5pt, right=5pt, top=4pt, bottom=4pt,
  fonttitle=\bfseries\footnotesize, fontupper=\footnotesize,
  before upper={\raggedright\setlength{\parindent}{0pt}}}}
\newlist{pitems}{itemize}{2}
\setlist[pitems]{label=\textbullet, leftmargin=1.1em, itemsep=0.5pt, topsep=1.5pt, parsep=0pt}
\definecolor{mypurple}{HTML}{56517E}
\definecolor{mygreen}{HTML}{4D9900}
\definecolor{myred}{HTML}{F02400}
\definecolor{mygold}{HTML}{D79B00}
\definecolor{myblue}{HTML}{6C8EBF}
\definecolor{royalblue}{HTML}{0066CC}
\definecolor{myturq}{HTML}{0E8088}
\definecolor{mynavy}{HTML}{10739E}

\usepackage{array}
\newcommand{\term}[2]{\textbf{\begin{tabular}[t]{@{}l@{}}#1\\#2\end{tabular}}}
\usepackage{booktabs}

\usepackage{caption}
\usepackage{hyperref}

\newcommand{\tightparagraph}[1]{%
  \vspace{0.1em}
  \noindent\textbf{#1}%
}

\newenvironment{tightequation}{%
  \begingroup
  \footnotesize
  \setlength{\abovedisplayskip}{0pt}
  \setlength{\belowdisplayskip}{2pt}
  \setlength{\abovedisplayshortskip}{0pt}
  \setlength{\belowdisplayshortskip}{2pt}
}{%
  \endgroup
}
\newenvironment{tightdefinition}
  {%
   \begingroup
   \setlength{\abovedisplayskip}{1pt}%
   \setlength{\belowdisplayskip}{4pt}%
   \setlength{\abovedisplayshortskip}{1pt}%
   \setlength{\belowdisplayshortskip}{4pt}%
   \begin{equation}%
  }
  {%
   \end{equation}%
   \endgroup
  }

\usepackage{float}
\usepackage{placeins}
\newcommand{\llmr}{\textsc{LLM-Refiner}}
\newcommand{\llmrs}{\textsc{LLM-R}}

\newcommand{\graphoverlap}{\hyperlink{term:graphical-overlap}{\textbf{Local Graph Overlap}}}
\newcommand{\semoverlap}{\hyperlink{term:semantic-overlap}{\textbf{Semantic Overlap}}}

\newcommand{\semneighborhood}{\hyperlink{term:semantic-neighborhood}{\textit{Semantic Neighborhood}}}
\newcommand{\semkeystone}{\hyperlink{term:semantic-keystone}{\textit{Semantic Keystone}}}
\newcommand{\semintake}{\hyperlink{term:semantic-intake}{\textsc{Semantic Intake}}}

\newcommand{\lingcontext}{\hyperlink{term:linguistic-context}{\textit{Linguistic Context}}}

\newcommand{\graphoverlapP}{\textbf{Graphical Overlap}}
\newcommand{\semoverlapP}{\textbf{Semantic Overlap}}

\newcommand{\semneighborhoodP}{\textit{Semantic Neighborhood}}
\newcommand{\semkeystoneP}{\textit{Semantic Keystone}}
\newcommand{\semintakeP}{\textsc{Semantic Intake}}

\newcommand{\stpP}{\textsc{Semantic Test Protocol}}
\newcommand{\gtpP}{\textit{Graphical Test Protocol}}
\newcommand{\lingcontextP}{\textit{Linguistic Context}}

\title{Putting Meaning Back in Abstract Meaning Representation: \\ Using LLMs to Repair Semantic Errors of AMR Parsers}

\author{Ian Yuen \\
  Independent Researcher \\
  \texttt{jardinianr@gmail.com} \\\And
  Douglas S. Chan \\
  Cayuga Wireless \\
  \texttt{douglas.chan@ieee.org} \\\And
  Nikita Mishra \\
  Amazon.com \\
  \texttt{mishniki@amazon.com} \\}

\begin{document}
\maketitle
\begin{abstract}
AMR parsers now approach human-level SMATCH scores. Yet state-of-the-art parsers may still omit or distort critical sentence semantics, causing high amounts of local graph overlap that misrepresent global meaning errors. We address this \textit{residual semantic gap} with our \llmr{} (\llmrs{}), a post-editing pipeline that diagnoses semantic errors in strong parser outputs and applies localized AMR repairs. We introduce a \stpP{} to better evaluate whether a repaired AMR has preserved a sentence's meaning. On GrAPES dataset, \llmrs{} improves AMRBart performance by +5.2\% in SemBLEU and +16\% on the subset that \llmrs{} repaired. Our analysis also reveals SMATCH understates many meaning-preserving repairs captured by SemBLEU. These results suggest the next frontier of AMR parsing should optimize not only local graph overlaps, but also global semantic overlaps. This can open a potential path to reliable \textit{automatic} AMR parsers, which could facilitate more symbolic datasets and unblock neuro-symbolic research at scale.
\end{abstract}

\section{Introduction}
Abstract Meaning Representation (AMR) uses a graph to represent a sentence's meaning \citep{banarescu-etal-2013-abstract}. Because \textit{gold AMR} graphs (i.e., ground-truths) require human expert annotation, AMR datasets are expensive to scale. 
\textit{Automatic} text-to-AMR parsing is therefore critical to expanding semantic annotation and support large-scale training of neuro-symbolic systems. Improving neuro-symbolic methods could potentially address interpretability and hallucinations issues in current models as well \citep{agrawal-etal-2024-knowledge}. Ultimately, advancing state-of-the-art AMR parsing will have broad benefits for downstream NLP applications \citep{wein-opitz-2024-survey,li-etal-2025-sentence,michail-etal-2026-alee,ye-etal-2026-swan}.

Recent AMR parsers have advanced significantly to attain human inter-annotator-agreement (\textit{IAA}) \citep{banarescu-etal-2013-abstract} level. In this paper, we define a \textit{strong} parser, like \mbox{AMRBart} \citep{bai-etal-2022-graph}, as one that surpasses an \textit{IAA} of $0.83$ for SMATCH, the de facto graph metric for AMR parsing.

\begin{figure}[t]
    \centering
    \includegraphics[width=\columnwidth]{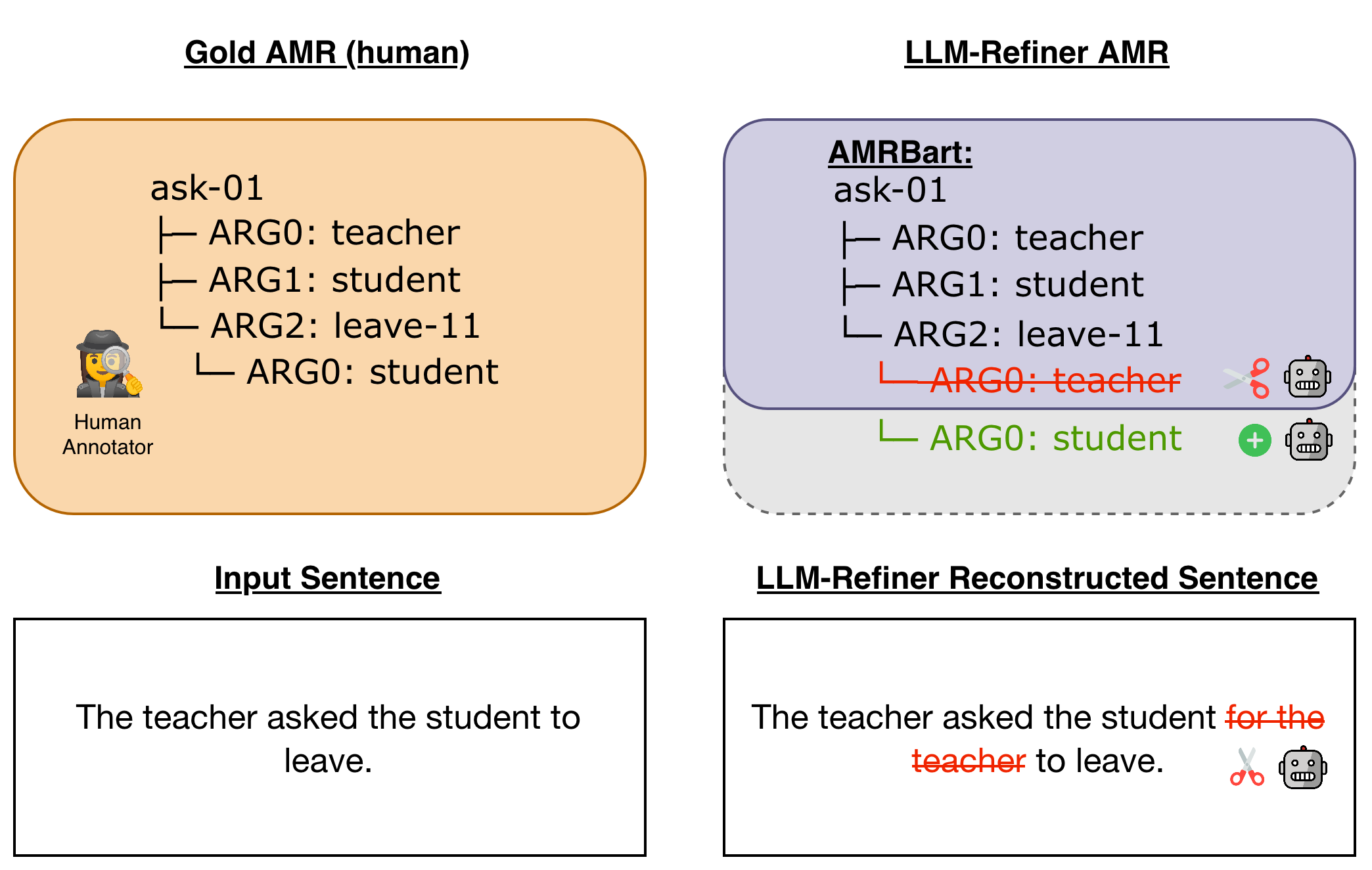}
    \caption{\llmr{} corrected the subject–object relation of the ``leave'' action (\textit{leave-11}), attaching the edge to the correct subject, \textcolor{mygreen}{student}, and removing the edge to the incorrect subject \textcolor{myred}{teacher}.  SemBLEU is more sensitive than SMATCH at detecting this type of \textit{residual semantic gap} and accounting for its improvement (see Sec~.\ref{sec:sembleu-adv}).}
    \label{fig:sem}
\end{figure}

But a high SMATCH-scoring AMR graph that closely resembles the gold graph can still err on essential concepts, like "who did what to whom" \citep{opitz-frank-2022-better}. These errors can produce large global semantic impacts not reflected in SMATCH, which merits on degree of local graph overlap (i.e., resemblance between node-edge-node \textit{bi-grams}). We denote these failure phenomena as \textit{residual semantic gaps} in a parser's output (see Sec.~\ref{sec:sembleu} for details).

To target these gaps, we introduce \llmr{} (\llmrs{}), an LLM-aided post-editor pipeline that extends an existing AMR parser. \llmrs{} starts from a strong parser's AMR output, diagnoses its missing or distorted sentence-level meanings via prompting an LLM, and repairs these errors.

Note that improvements from \llmrs{}'s correct repairs may not be captured by SMATCH.  This is due to a core assumption in SMATCH's AMR evaluation: A high score corresponds to \textit{how closely a predicted graph matches the exact local}  \textit{bi-gram} \textit{(node-edge-node) structures of a human labeled} \textit{gold} \textit{AMR}.  We define this measure as \graphoverlap{} below. Although useful in many cases, it fails to recognize a graph may be expressing the wrong meaning globally while preserving many local \textit{gold} structures, thus attaining an erroneously high SMATCH.  Conversely, a small local repair can significantly improve global meaning representation, this can be seen in Fig.~\ref{fig:sem}. Evidently, \textbf{Local Graph Overlap} is not a sufficiently accurate measure for the final output's semantic gaps. 

We therefore propose the \textit{semantic dimension} as a new frontier for AMR parser performance (detailed in Sec~\ref{sec:taxonomy}). Accordingly, we define: 
\semoverlap{} that measures \textit{whether the predicted AMR contains graphical structure that express local and global meaning correctly}.
This semantic dimension captures the \textit{residual semantic gap} that SMATCH can omit; this is detrimental because there are local graph errors that change the overall graph's meaning.

Altogether, our contributions in this paper are:
\begin{itemize}[itemsep=2pt,topsep=3pt,leftmargin=*]
    \item We propose a taxonomy to describe the semantic dimensions for improving AMR performance and evaluation (Table~\ref{tab:taxonomy}).
    \item We introduce \llmr{}, a multi-LLM-models pipeline to post-edit semantic errors of strong AMR parser outputs. \llmrs{} improves AMRBart outputs on GrAPES by +5.2\% in SemBLEU and +16\% on the subset that \llmrs{} edited.
    \item We demonstrate that augmenting \textsc{LLM-R} with \hyperlink{term:linguistic-context}{\textit{Linguistic Context}} yields +41\% improvement relative to without.
    \item We propose the \stpP{} (GrAPES; SemBLEU) over the standard \gtpP{} (AMR3; SMATCH) for evaluating semantic repairs and discuss empirical results demonstrating its relative effectiveness.
\end{itemize}

\section{Related Work}

\label{sec:related}

\tightparagraph{\textbf{Semantic Gaps in AMR Parsers:}} Researchers have explored residual semantic gaps not reflected by \textit{IAA}-level SMATCH scores. \citet{opitz-frank-2022-better} showed that strong parsers, like AMRBart, still produce meaning-level failures despite high SMATCH.  Subsequent work traced these errors to recurring linguistic phenomena such as reentrancies, missing edges, and relative-clause structures \citep{szubert-etal-2020-role,yang-schneider-2024-relative}. \citet{groschwitz-etal-2023-amr} operationalized this concern in GrAPES, a diagnostic dataset targeting linguistic phenomena that the broad-coverage AMR3 dataset obscure. Our Sec.~\ref{sec:grapes} is among the first to thoroughly compare GrAPES and AMR3.

\tightparagraph{\textbf{Redesigning AMR Evaluation Metrics:}} Residual semantic errors have motivated evaluation metrics beyond SMATCH. SemBLEU \citep{song-gildea-2019-sembleu} compares multi-hop graph {\textit{$k$-grams}, whereas SMATCH primarily rewards overlap of single-hop graph \textit{bi-grams}. This makes SemBLEU more sensitive to local compositional context: a node must appear with the correct surrounding structure, not just correct in an isolated matching edge (see Sec.~\ref{sec:sembleu-adv} for discussion). There are other metrics that also augment some graph overlap with semantic similarity or structure-aware comparison, but they are not as sensitive as SemBLEU \citep{opitz-frank-2022-better}.

\tightparagraph{\textbf{LLM as AMR Parser:}} Newer LMs and LLMs have not solved AMR parsing directly. Recent systems using OpenAI GPT-4o, Meta LLaMA-GRPO, and Flan-T5 still trail AMRBart on AMR3 (see Table~\ref{tab:metric-results}) \citep{li-fowlie-2025-gpt,barta-etal-2025-enhancing,Lee2023AMRParsingInstructionFinetuned}. Moreover, LLM-generated parses can contain critical semantic inaccuracies \citep{ettinger-etal-2023-expert}. This presents a paradox: newer LM models with stronger general language abilities underperform older sequence-to-sequence parsers. We initiate an investigation into this by separating linguistic reasoning and AMR formalization within a multi-model pipeline.

\tightparagraph{\textbf{LLM as Post-Editor:}} LLMs as post-editors have shown successes in adjacent fields, like machine translation. Metrics of implementations with LLM post-editors exceeded those that generate outputs from scratch \citep{raunak-etal-2023-leveraging,ki-carpuat-2024-guiding}. To our knowledge, we are the first to introduce LLM post-editors in AMR parsing.

\tightparagraph{\textbf{LLMs with Task Context:}}
\label{para:llm-task-context}
Researchers have shown that domain-specific context \citep{Xu2023ExpertPrompting}, role specification \citep{kong-etal-2024-better}, and domain-specific label spaces \citep{min-etal-2022-rethinking} can improve downstream performance in LLM generation. Our pipeline's post-editor design is a type of task context LLM (Sec.\ref{sec:pipeline}).

\tightparagraph{\textbf{Structured LLM Inference:}}
A related technique in LLM design we also incorporated in our pipeline is decomposing output generation into explicit stages: proposing, comparing, ranking, and filtering candidate responses. These structured inference procedures reduce reliance on a single unconstrained generation and improve downstream task performance \citep{liu-etal-2024-longgenbench, hu-etal-2024-wavllm}.

\tightparagraph{\textbf{Neuro-symbolic Performance:}} Recent neuro-symbolic research shows that symbolic structure can improve neural networks. AMR is a natural symbolic interface for language modeling. Incorporating it into pretrained models --- like, through GNNs, fine-tuning, and AMR-aware tokenization --- can improve text-only baselines on question answering, classification, summarization, and knowledge-intensive tasks \citep{hua-etal-2022-amrtvsumm, li-etal-2025-injecting, pham-etal-2024-graph, wang-etal-2023-exploiting, shou-etal-2022-amr}. These AMR-enhanced NLP results show paths towards the broader neural-symbolic goal of combining learned representations with explicit symbolic structure \citep{valiant-2008-knowledge-infusion, garcez-etal-2009-neural-symbolic, garcez-etal-2015-neural-symbolic}. However, applying AMR at scale depends on reliable automatic parsing rather than via small manually annotated datasets as discussed in these prior works. Therefore, this paper's research is critical as it may be a path towards generalizing neuro-symbolic gains in realistic settings.

\begin{figure*}[t!]
    \centering
    \includegraphics[width=\textwidth]{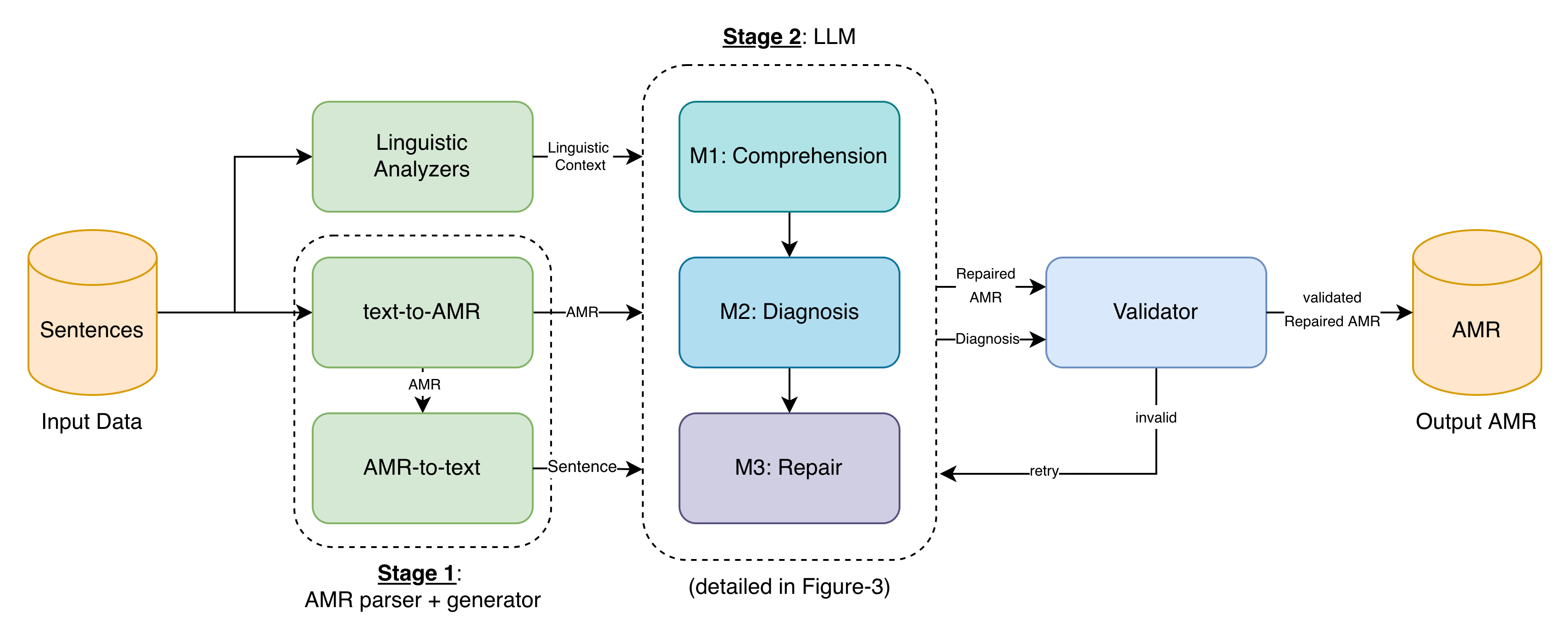}
    \caption{\llmr{} post-edits output from AMRBart with a multi-stage approach, separating LLM semantic reasoning (\textcolor{myturq}{M1: Comprehension}, \textcolor{mynavy}{M2: Diagnosis}) from AMR formalization (\textcolor{mypurple}{M3: Repair}), detailed in Fig~\ref{fig:module}.}
    \label{fig:overview}
\end{figure*}

\begin{table}[t]
\centering
\footnotesize
\setlength{\tabcolsep}{4pt}
\renewcommand{\arraystretch}{1.22}
\begin{tabular}{p{0.3\linewidth}p{0.64\linewidth}}
\hline
\textbf{Term} & \textbf{Definition} \\
\hline

\raisebox{1.2\baselineskip}[0pt][0pt]{%
  \hypertarget{term:semantic-neighborhood}{}%
}%
\term{Semantic}{Neighborhood} &
Node and edge sequences connected by bi-grams (one-hop) and k-grams (k-hops) in an AMR graph. \\
\arrayrulecolor{gray!25}\hline\arrayrulecolor{black}

\raisebox{1.2\baselineskip}[0pt][0pt]{%
  \hypertarget{term:semantic-keystone}{}%
}%
\term{Semantic}{Keystone} &
A node / edge / subgraph contained in high proportion of \semneighborhoodP{}s (high node centrality -- i.e., the node is highly connected with other nodes), which when repaired, causes a large delta in meaning, be it better or worse. \\
\arrayrulecolor{gray!25}\hline\arrayrulecolor{black}

\noalign{\hrule height 1pt}

\raisebox{1.2\baselineskip}[0pt][0pt]{%
  \hypertarget{term:grahical-overlap}{}%
}%
\term{Local}{Graph Overlap} &
Evaluates AMR by counting one-hop \semneighborhoodP{} (bi-grams) shared between output and gold AMR, e.g., SMATCH. \\
\arrayrulecolor{gray!25}\hline\arrayrulecolor{black}

\raisebox{1.2\baselineskip}[0pt][0pt]{%
  \hypertarget{term:semantic-overlap}{}%
}%
\term{Semantic}{Overlap} &
Evaluates AMR by counting $k$-hop \semneighborhoodP{}s (k-grams) ($k=1,\ldots,K$) shared between output and gold AMR, e.g., SemBLEU. \\
\arrayrulecolor{gray!25}\hline\arrayrulecolor{black}

\noalign{\hrule height 1pt}

\raisebox{1.2\baselineskip}[0pt][0pt]{%
  \hypertarget{term:semantic-intake}{}%
}%
\term{Semantic}{Intake} &
Semantic categories from linguistic theory used to decompose a sentence into semantic components. (See Table~\ref{tab:appendix-linguistic-categories}).  \\
\arrayrulecolor{gray!25}\hline\arrayrulecolor{black}

\raisebox{1.2\baselineskip}[0pt][0pt]{%
  \hypertarget{term:linguistic-context}{}%
}%
\term{Linguistic}{Context} &
Linguistic information derived from input sentence with open-source NLP tools, e.g., POS tags, dependency parses. (See \ref{app:linguistic-context} for details.)\\

\noalign{\hrule height 1pt}

\raisebox{1.2\baselineskip}[0pt][0pt]{%
  \hypertarget{term:semantic-test-protocol}{}%
}%
\term{Semantic}{Test Protocol} &
A stress-test setting for semantic errors in AMR: \{GrAPES; SemBLEU\} (See Sec.~\ref{sec:sts} for details). \\
\arrayrulecolor{gray!25}\hline\arrayrulecolor{black}

\raisebox{1.2\baselineskip}[0pt][0pt]{%
  \hypertarget{term:graphical-test-protocol}{}%
}%
\term{Graphical}{Test Protocol} &
The current standard \textbf{Local Graph Overlap} setting for AMR evaluation: \{AMR3; SMATCH\}. \\
\arrayrulecolor{gray!25}\hline\arrayrulecolor{black}

\noalign{\hrule height 1pt}

\hline
\end{tabular}
\caption{A taxonomy for describing the \textit{semantic dimension} in AMR.}
\label{tab:taxonomy}
\end{table}

\section{Semantic Taxonomy for AMR Research}
\label{sec:taxonomy}

Historically, the development of AMR parsers have been heavily influenced by SMATCH. Because the incentive is to maximize for SMATCH scores, the accuracy of an AMR graph is measured by its amount of local overlap with the gold graph based on graphical concepts --- like variable alignment,  \textit{bi-gram} overlap, and node/edge recall.

While this approach can facilitate scalable evaluation, local graphical overlaps alone do not provide any indications of the semantic content represented by the graph. In other words, this \textit{semantic dimension} of a graph is still inadequately represented in current AMR taxonomy and should be formalized explicitly.

The disconnect between structural graph and semantic graph concepts is also echoed in early Semantic Web research. Their foundational Resource Description Framework is a taxonomy of semantic concepts overlaid on graphical concepts.  But structural graph equivalence does not carry over to semantic equivalence \citep{Gutierrez2004Foundations}. Similarly, AMR metric research has shown high SMATCH scores that focus on local graphical overlap can still obfuscate meaning-distorting errors \citep{opitz-frank-2022-better}.

Based on these motivating factors, we propose a foundational taxonomy in Table~\ref{tab:taxonomy} to explicitly define the semantic dimension of a given AMR graph. This can provide an optimization target properly rooted on semantic meanings for furthering research of AMR datasets, parsers, and evaluation.

\section{Semantic Test Protocol (STP)}
\label{sec:sts}

Commonly reported datasets and metrics for AMR evaluation primarily measure graph overlap which can overlook the \textit{semantic dimension}.  Although recent AMR research has proposed new tools to address perceived gaps, they are not widely adopted and applied mutually exclusively.  With Sec.~\ref{sec:taxonomy}’s \textit{Semantic Taxonomy}, we can now better distinguish datasets and metrics that reflect more \textit{semantic dimensions}.  Based on that, we propose below the \stpP{} and its dataset and metric. We explain how STP, which targets semantic overlap, is complementary to current tools that target local graphical overlap. We denote the latter tools as \gtpP{} (GTP) and provide its definition in Table~\ref{tab:taxonomy}. As STP and GTP provide separately useful signals of AMR accuracy, the two should be used together and not mutually exclusively. This is especially important as the residual gaps in state-of-the-art AMR parsers are simultaneously graphical and semantic issues; fixing this is the goal of this paper. 

We present in Sec.~\ref{sec:results} empirical results of applying STP to evaluate AMR parsers and show in Sec.~\ref{sec:sembleu-adv} that the results validate STP’s efficacy in spotlighting the residual semantic gaps. Having a means to identify them can aid the design of AMR parsers --- like how we do so with our \llmr{} (in Sec.~\ref{sec:pipeline}) to target these gaps when repairing strong AMR parsers’ outputs.

\subsection{STP Dataset: GrAPES}
\label{sec:grapes}
Our STP’s evaluation dataset is GrAPES \citep{groschwitz-etal-2023-amr} --- a diagnostic benchmark built around semantically complex sentences and known AMR parser failure modes (see examples from literature in Table~\ref{tab:appendix-synthetic-subcorpora}). The commonly adapted AMR3 dataset, by contrast, is a broad-coverage corpus drawn from newspapers and web sources (Table~\ref{tab:appendix-source-datasets}), and often entangled with metadata-like artifacts (Table~\ref{tab:appendix-amr3-nondata}).

As a result, AMR3 is useful for measuring aggregate parser quality, but its sources mostly reflect everyday language, which has sparse semantically complex phenomena, and thus lead to less residual semantic errors in parser output, making these residual errors and improvements on it harder to detect. 
Therefore, GrAPES is a sharper test of residual semantic gaps.

\subsection{STP Metrics: SemBLEU}
\label{sec:sembleu}

For evaluation metrics, our STP uses SemBLEU \citep{song-gildea-2019-sembleu} because it is designed to measure semantic repairs that SMATCH can understate. SMATCH is restricted to graphical overlap only between \textit{bi-grams} of a parser output and \textit{gold} \citep{cai-knight-2013-smatch}; whereas SemBLEU compares \textit{k-grams} for various $k$ \citep{song-gildea-2019-sembleu}. Thus, for a given node or edge, SemBLEU counts its higher-order \semneighborhoodP{}s, which correspond to a larger semantic context; whilst SMATCH counts only the first order (see Sec.~\ref{sec:sembleu-adv} for detailed discussion). 

We define $\Delta \text{metric}$ as the change in a metric score after \llmr{} repairs an AMR; i.e., for each sample, we compare both the repaired AMR and the parser's initial AMR against the \textit{gold} AMR, and take the difference between those two scores. Hence, $\Delta \mathrm{SemBLEU}$ is our primary metric for the repair of residual semantic gap in strong parsers.

\begin{figure*}[t]
    \centering
    \includegraphics[width=0.7\textwidth]{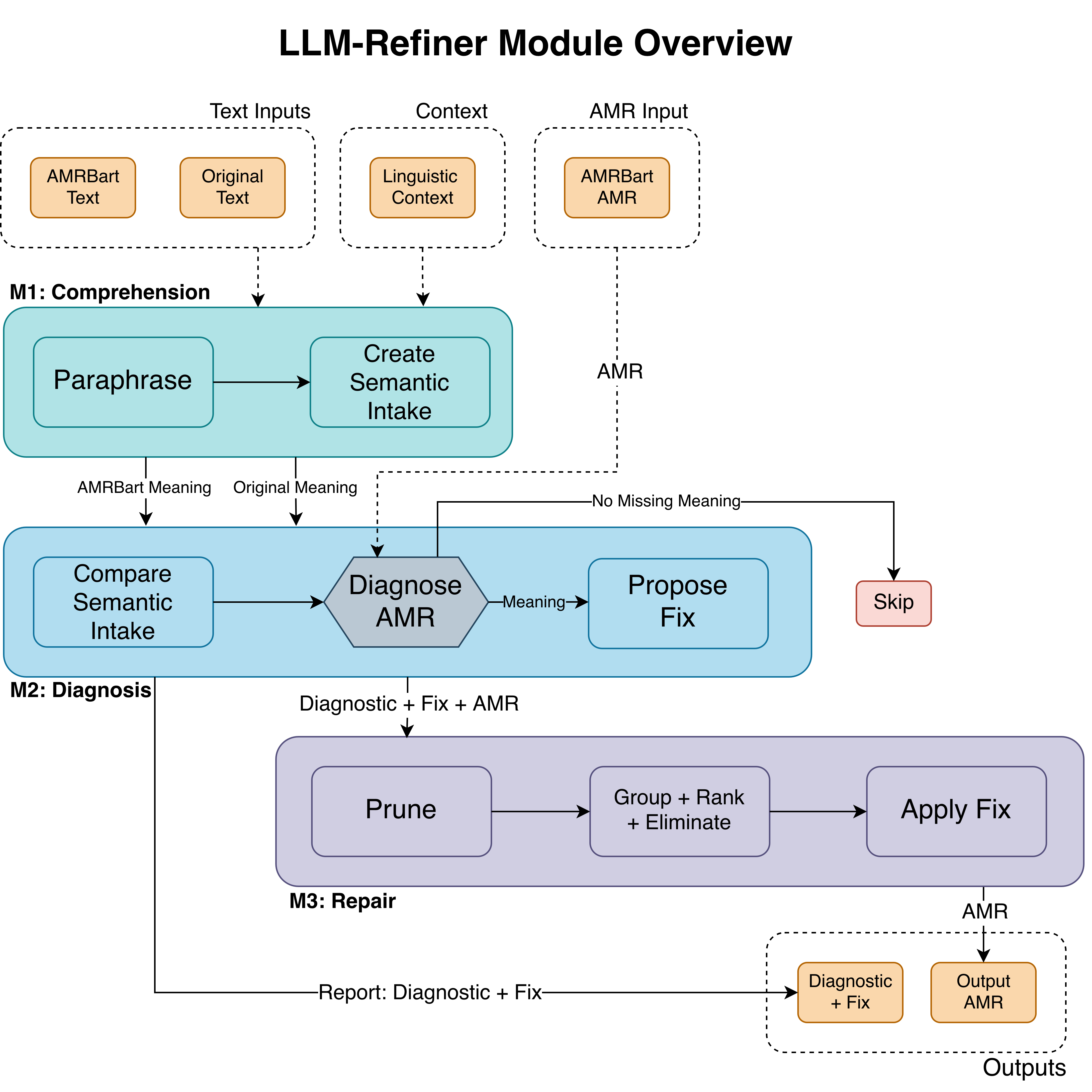}
    \caption{Stage 2, \llmr{} organizes processing across modules with \semintakeP{} as the intermediate representation of meaning. It outputs the completed intake with the diagnosis and repaired AMR.}
    \label{fig:module}
\end{figure*}

\section{\llmr{} Pipeline}
\label{sec:pipeline}
Our \llmr{} is a multi-model pipeline, where models process input in sequential stages (see Fig.~\ref{fig:overview}).  We describe below how \llmrs{} targets the residual semantic gap exhibited in strong AMR parsers in terms of Sec.~\ref{sec:taxonomy}'s \textit{Semantic Taxonomy}.

In Stage 1, an AMR parser-and-generator module produces an initial AMR and reconstructed sentence, respectively via text-to-AMR and AMR-to-text.  This module can be AMRBart. By starting from a parser's AMR output as opposed to an input sentence, we constrain the LLM's task to post-editing and restrict hallucinations.

In Stage 2, we invoke LLM and embed the semantic dimension into its prompt. LLM (e.g. GPT-4o) is directed to diagnose and fix semantic gaps in three steps (Fig~\ref{fig:module}). \textcolor{myturq}{In M1}, LLM decomposes the input and reconstructed sentences into semantic units by linguistic categories in \semintakeP{}  (Box~\ref{box:step1-reconstruction} in \ref{app:prompts}).  They are compiled from formal semantics research and diagnostic datasets \citep{poliak-2020-survey,Mao2023}. These semantic units are analogous to those that \semoverlapP{} operates on. This guardrail keeps LLM's reasoning within linguistic theory and enables cross-stage modularity and interpretability.

Then \textcolor{mynavy}{M2} looks for semantic gaps in each sample, using the gap categories of Table~\ref{tab:appendix-linguistic-categories}. The LLM compares the \semintakeP{} between input and reconstructed sentence row by row (Box~\ref{box:m2-diagnosis} in \ref{app:prompts}). For each gap identified, a suggested AMR repair is made and the corresponding statement of diagnosis is appended (hexagon in Fig~\ref{fig:module}). This restricts LLM to point-wise comparison and makes the diagnosis process human interpretable. 

In \textcolor{mypurple}{M3}, all rows are consolidated, confidence scored, ranked, filtered, and with remaining AMR edits applied. This controls over-generation and hallucinations. To enforce graph validity, the Validator module compiles the AMR return string output (\ref{app:json-schema}) from Stage 2 into Penman graph format, via the Python Penman library \citep{goodman-2020-penman}.  It throws an error when the string's AMR format is wrong; the Validator then returns the string back to Stage 2 and prompts LLM to fix any invalid AMR formatting.

\llmrs{} is designed to be model agnostic, with conditions for practicality: (1) Stage 1 models must produce high quality AMR with only residual errors for post-editing, proxied as reaching \textit{IAA} SMATCH levels (discussed further in Sec.~\ref{sec:limits}). Accordingly, we report results for these strong AMR parser models: AMRBart and SPRING. (2) Stage 2 requires an LLM size that balances smaller models' under-generation of necessary repairs against larger models' over-generation of extraneous repairs (Sec.~\ref{sec:ablations}). We found \llmrs{} to perform best with GPT-4o, which is also the more widely reported LLM in AMR literature.

\vspace{0.2em}

\tightparagraph{\textbf{LLM Task Context with \lingcontext{} -- Linguistic Analyzers}}: To further investigate LLM improvements with increased context, we attach \lingcontextP{} generated from input sentence with open-source \textit{Linguistic Analyzers} (listed in \ref{app:linguistic-context}).  As this provides deterministic anchors for the LLM to reason over, linguistic reasoning load is thus reduced. We refer to \llmrs{} with \lingcontextP{} as \textsc{LLM-R+L}.

\vspace{0.2em}

\tightparagraph{\textbf{Highlighting \llmr{}'s Repair Gains}}:
\label{sec:complex} To clearly demonstrate gains from \llmrs{}'s repairs, we also focus on the subset of samples that M2 diagnosed as needing repairs.  But because LLM diagnosis is probabilistic, such a subset would be different across runs.  So, we examine the \textit{Repair Subset}, $R$, that is the intersection of such a subset across three different runs of a sentence being examined.\footnote{While we can find $R$ manually, to make this scalable, we implement \semintake{} in the diagnosis stage of \llmrs{} and separate it from the repair stage.}  In other words, $R$ contains samples that \llmr{} identified to have semantic gaps in all three runs.

In Sec.~\ref{sec:results} we report $R$ specific performance as an auxiliary metric. The three subsets here overlap by approximately 82\% and each subset's size as a proportion of the full dataset remains stable.

\begin{table*}[!htbp]
\centering
\small
\setlength{\tabcolsep}{4.5pt}
\begin{tabular}{lc|cccc|c|cccc}
\hline
\textbf{Dataset} &
\multicolumn{5}{c|}{\textbf{SemBLEU}} &
\multicolumn{5}{c}{\textbf{SMATCH}} \\
\cline{2-6} \cline{7-11}
& AMRBart & LLM-R & $\Delta$\% & LLM-R+L & $\Delta$\%
& AMRBart & LLM-R & $\Delta$\% & LLM-R+L & $\Delta$\% \\
\hline
\noalign{\vskip 2pt}
GrAPES          
& 0.671 & 0.695 & +3.7\%  & 0.706 & +5.2\% 
& 0.800 & 0.815 & +1.9\% & 0.822 & +2.7\% \\

GrAPES$^R$
& 0.639 & 0.746 & +16\%   & 0.780 & +22\% 
& 0.788 & 0.854 & +8.4\% & 0.876 & +11\% \\

AMR3          
& 0.721 & 0.724 & +0.5\%  & 0.728 & +0.9\% 
& \textbf{0.842} & 0.843 & +0.2\% & 0.845 & +0.4\% \\

AMR3$^R$
& 0.699 & 0.719 & +2.9\%  & 0.735 & +5.1\% 
& 0.807 & 0.813 & +0.7\% & 0.823 & +1.9\% \\
\hline
\end{tabular}
\caption{SemBLEU and SMATCH results comparing AMRBart, \textsc{LLM-R+L} post-editing, and \textsc{LLM-R} variants. Previous work models trail AMRBart and only report \textbf{SMATCH on AMR3} (GPT-4o: \textbf{0.60}, LLaMA-GRPO: \textbf{0.80}, Flan-T5: \textbf{0.83}, SPRING: \textbf{0.83}). For SPRING (Table~\ref{tab:metric-results-spring}) gains are muted on AMR3, where both parsers are near IAA, and smaller on GrAPES for SPRING because it starts from a stronger baseline. All results are statistically significant (CI reported in Table~\ref{tab:metric-results-ci}).}
\label{tab:metric-results}
\end{table*}

\begin{table}[t]
    \centering
    \footnotesize
    \begin{tabular}{lccc}
        \hline
        \textbf{Metric} & \textbf{AMRBart} & \textbf{LLM-R} & \textbf{$\Delta$\%} \\
        \hline
        \noalign{\vskip 2pt}
        s2match & 0.798 & 0.863 & +8.10\% \\
        wlk & 0.768 & 0.796 & +3.68\% \\
        wwlk\_k2 & 0.822 & 0.853 & +3.76\% \\
        wwlk\_k3e2n & 0.812 & 0.828 & +1.84\% \\
        \hline
    \end{tabular}
    \caption{\llmr{} improvements reflected in other AMR evaluation metrics \citep{opitz-frank-2022-better}.}
    \label{tab:add-metric-results}
\end{table}

\section{\llmr{} Performance Results}
\label{sec:results}

We evaluate \llmr{}’s performance with our STP and show them in Table~\ref{tab:metric-results} alongside GTP results. For SemBLEU metrics, \llmr{} with \lingcontextP{} (\textsc{LLM-R+L}) improved on AMRBart by 5.2\% for the GrAPES dataset and 16\% on its \textit{repair subset} $R$ (denoted by GrAPES$^R$). SMATCH mirrors same trends: gains of 2.7\% on GrAPES and 11\% on GrAPES$^R$.  These illustrate sensitivity differences between SemBLEU and SMATCH, discussed in Sec.~\ref{sec:sembleu-adv}.  \llmrs{}'s SMATCH gains on AMR3 are more pronounced when examining its $R$ (denoted by AMR3$^R$) at 2.9\%. This is expected as AMRBart training specialized on AMR3 for achieving high SMATCH (Sec.~\ref{sec:ablations}). Overall, \llmrs{} exhibits similar gains across other AMR evaluation metrics (Table~\ref{tab:add-metric-results}).

On average, \textsc{LLM-R+L} identifies and repairs samples with residual semantic gap for 31\% of GrAPES and 6\% of AMR3 for each respective dataset (stable $\pm3\%$ across runs) (\ref{app:repairs}). These fixes align with failure modes identified in previous research (Sec.\ref{sec:related}). (See Table~\ref{tab:refiner-categories} and Table~\ref{tab:appendix-linguistic-categories} for details.) This shows that \llmr{} can effectively address the \textit{residual semantic gaps} in AMRBart \footnote{Manual evaluation of one run showed 91\% accuracy.}.

As AMRBart had already reached \textit{IAA}, only a small ``residual'' proportion of samples are incorrect and need fixing \citep{opitz-frank-2022-better}.  But repairing them could improve the accuracy significantly --- this effect is visible with the GrAPES$^R$ and AMR3$^R$ rows of Table~\ref{tab:metric-results}. These rows represent samples that have semantic gaps and would include residual portions that AMRBart had missed.

\tightparagraph{\textbf{Linguistic Context:}} Relative gains by \textsc{LLM-R} over \textsc{LLM-R+L} in Table~\ref{tab:metric-results} confirm benefits from \textit{Linguistic Analyzers} of Sec.~\ref{sec:pipeline}. E.g., \textsc{LLM-R+L} has a 41\% relative gain on GrAPES for SemBLEU.

\vspace{0.2em}

\begin{table*}[!htbp]
\centering
\small
\setlength{\tabcolsep}{4.5pt}
\begin{tabular}{lc|cccc|c|cccc}
\hline
\textbf{Dataset} &
\multicolumn{5}{c|}{\textbf{SemBLEU}} &
\multicolumn{5}{c}{\textbf{SMATCH}} \\
\cline{2-6} \cline{7-11}
& SPRING & LLM-R & $\Delta$\% & LLM-R+L & $\Delta$\%
& SPRING & LLM-R & $\Delta$\% & LLM-R+L & $\Delta$\% \\
\hline
\noalign{\vskip 2pt}
GrAPES         
& 0.738 & 0.743 & +0.7\%  & 0.753 & +2.1\% 
& 0.835 & 0.841 & +0.8\% & 0.848 & +1.5\% \\

GrAPES$^R$
& 0.709 & 0.729 & +2.8\%   & 0.753 & +6.2\% 
& 0.827 & 0.842 & +1.8\% & 0.851 & +2.9\% \\

AMR3    
& 0.684 & 0.685 & +0.2\%  & 0.686 & +0.3\% 
& 0.828& 0.829 & +0.1\% & 0.830 & +0.2\% \\

AMR3$^R$
& 0.645 & 0.656 & +1.8\%  & 0.661 & +2.5\% 
& 0.801 & 0.803 & +0.3\% & 0.810 & +1.1\% \\
\hline
\end{tabular}
\caption{SemBLEU and SMATCH results comparing SPRING, \textsc{LLM-R+L} post-editing, and \textsc{LLM-R} variants. Gains are muted on AMR3, where both parsers are near \textit{IAA}, and smaller on GrAPES for SPRING because it starts from a stronger baseline. All results are statistically significant.}
\label{tab:metric-results-spring}
\end{table*}

\section{Discussion on Discrepancy between STP and GTP Metrics}
\label{sec:sembleu-adv}

We observed that \llmrs{}'s SMATCH metrics in Table 2 do not appear to show proportionate gains as SemBLEU does.  Motivated by this inconsistency conundrum, we analyze the reported metrics (SemBLEU and SMATCH) and confirmed the reasons are related to the overlooked \textit{semantic dimension} in SMATCH.  This is evident from how SemBLEU is more sensitive in detecting residual semantic gaps in AMR parsers, for which we explain below.  Although \llmr{} fixes these gaps, SMATCH’s metric fails to fully account for them.

To identify this discrepancy between SemBLEU and SMATCH, we introduce the concept of \semneighborhoodP{}, a term in Sec.~\ref{sec:taxonomy} \textit{Semantic Taxonomy}. SemBLEU approaches \semneighborhoodP{} through \semoverlapP{}, while SMATCH approaches through \graphoverlapP{}.

Referencing \citet{song-gildea-2019-sembleu}'s definition of \textit{k-gram} overlap for $k=1,\ldots,K$, the \textit{k-gram} counting mechanism in SemBLEU and SMATCH is

\begin{nolinenumbers}
\begin{tightdefinition}
\label{eq:kgram}
p_k(c,z)
=
\frac{
\left|
\mathrm{kgram}(z)
\cap
\mathrm{kgram}(c)
\right|
}{
\left|
\mathrm{kgram}(c)
\right|
},
\end{tightdefinition}
\end{nolinenumbers}

\noindent where the numerator measures \textit{k-gram}s shared by graph $z$ and \textit{gold} graph $c$, and the denominator measures the total \textit{k-gram}s in $c$. We can interpret $p_k$  as the \semneighborhoodP{} which represents the semantic contexts to which a graph element belongs, at $k$ number of hops.

From this perspective, SMATCH only considers $p_2$ (\textit{bi-grams}), while SemBLEU can consider $\sum_{k=1}^{K}p_k$ and thus counts more \semneighborhoodP{}s than SMATCH. (See case study that shows this in \ref{sec:keystonecase}.)

As \llmr{} targets residual semantic gaps, its repairs tends to be on a \semkeystoneP{} (defined in Table~\ref{tab:taxonomy}).  Since \semkeystoneP{} is a highly connected node/edge that is contained in multiple \semneighborhoodP{}\textit{s}, this type of repairs will be semantically significant but has small structural graph changes.  Hence, 

\begin{tightequation}
\[
    \Delta \mathrm{SemBLEU}
    \gtrsim
    \Delta \mathrm{SMATCH}.
\]
\end{tightequation}

To observe this empirically, we can see in Table~\ref{tab:metric-results}, that for each dataset $\Delta \mathrm{SemBLEU}$ is on average 2x larger in value than $\Delta \mathrm{SMATCH}$ for the same repair. This suggests that SemBLEU is more sensitive to semantic gap repairs and supports the use of SemBLEU in our \stpP{}

\section{Ablation Analysis}
\label{sec:ablations}

Consistent with prior off-the-shelf LLM approaches to AMR parsing \citep{ettinger-etal-2023-expert,li-fowlie-2025-gpt}, we attempted a few-shot prompt that plainly asks the model to ``repair'' the AMR, but this led to unstable \llmr{} performance. For example, LLMs missed semantic errors, under-generated repairs, or introduced unsupported edits.
Ablating Stage 2 sub-modules (step-wise linguistic reasoning, linguistic context, or the confidence filter) produced similar instability, indicating that AMR post-editors require explicit semantic guardrails.

To test cross-parser generalization, we also swapped AMRBart with SPRING \citep{Bevilacqua2021SPRING} in the Stage 1 parser of \llmrs{}. Table~\ref{tab:metric-results-spring} shows SemBLEU gains of +2.1\% overall and +2.8\% on the GrAPES$^R$ subset, with weaker effects on AMR3 where residual errors are sparser. Gains are smaller than with AMRBart, possibly because AMRBart's fine-tuning on 200K SPRING-parsed Gigaword sentences \citep{rush-etal-2015-neural} had improved broad-coverage AMR3 performance while leaving more compositional errors still exposed on GrAPES. This contrast supports using GrAPES in general and the \stpP{} for evaluation, as it can surface gaps that broad-coverage benchmarks may obscure.

To test the modularity of \llmrs{} pipeline, we swapped Stage 1 parser and generator with other available models. We found that earlier rule-based and weaker sequence-based parsers at below \textit{IAA} produced output that are outside of \llmrs{}'s target of repairing residual semantic gaps, thus leading to Stage 2 LLMs producing incorrect repairs. Models with both text-to-AMR and AMR-to-text components, such as AMRBart and SPRING, performed best; mixing model families along these directions weakened repair accuracy.

We also substituted Stage 2 LLM with other available models. We found that smaller LLMs such as GPT-4o-mini often missed necessary repairs, while larger or reasoning-focused models such as GPT-5.2 and GPT-o3 tended to generate extraneous and unnecessary repairs. GPT-4o gave the best balance between being focused enough on small local repairs and being able to reason effectively through semantic nuances. As Claude Sonnet, Google Gemini, and xAI Grok showed weaker prompt transfer, so we selected to study a model from GPT-family, which are also more commonly reported in LLM-based AMR work.

Additionally, we conducted an ablation study using the open-source Google Gemma 4 31B model but observed no statistically significant improvements. Like smaller models, Gemma exhibited limitations in AMR construction. However, targeted prompt adjustments on GrAPES yielded promising gains --- we defer this to future research.

To test different architectures for targeting the residual semantic gap, we investigated different model design approaches. We found that multi-agent orchestration --- that with separate agents assigned to different linguistic categories --- increased cost, over-generated repairs, and struggled to preserve context across agents. We also attempted a purely rule-based repair engine, but found it to be limited because many residual AMR errors require context-dependent reasoning; as this requires complex linguistic rules which are hard to encode as deterministic rules.

\section{Discussion on Generalizability and Scalability}
\label{sec:general-scale}

\llmr{} inherits broad domain coverage from its Stage 1 AMR parser and Stage 2 LLM.
Empirically, this enables \llmr{} to operate well across both the mixed-genre examples in AMR3 and the compositional stress cases in GrAPES. Though, out-of-domain limitations remains; this is an active area of research (further details in Sec.~\ref{sec:limits}).

\llmr{} also improves the stability and semantic coherence of automatic AMR parser outputs, making parser-generated AMRs more viable as silver data for downstream AMR training.
Our pipeline is substantially faster than manual annotation: it produces a repaired AMR in roughly 15 seconds per sentence, compared with approximately 10 minutes for human annotation \citep{AMREditor}.
Furthermore, in our experiments, \llmrs{}'s average repair cost is below \$0.05 per sample.

\section{Conclusion}
\label{sec:conclusion}

We introduced \llmr{} that repairs the residual semantic gaps in strong AMR parsers. Our results motivated using our \stpP{} which more accurately illuminates AMR's semantic dimension.  We showed AMR parser and evaluation design should therefore ask not only whether a prediction matches locally, but whether it preserves the original meaning.

\section{Limitations}
\label{sec:limits}

This paper targets a specific weakness of current AMR parsers: repairing residual semantic errors that remain after strong AMR parsers already achieve high graph-overlap scores. We therefore did not exhaustively compare parsers, LLMs, prompts, or knowledge domains. Instead, our goal is to test whether \llmr{} and the \stpP{} can reveal and reduce semantic errors that standard graph metrics may under-report.

\llmr{} also inherits limitations in domain and model coverage from the datasets used by the parser being repaired. On AMR3, we showed \llmr{} can operate in the same broad-coverage setting as AMRBart and evidently \llmrs{} also attains or exceeds SMATCH \textit{IAA}. However, parser performance may degrade on specialized domains such as biomedical text \citep{bai-etal-2022-graph}; evaluating narrative fiction, Wikipedia, BioAMR, and other corpora remains future work. 

Additionally, \llmr{} relies on the availability of \textit{IAA} parsers trained on AMR datasets. To date, these datasets are primarily in English, while the open-source \textit{IAA} parsers are AMRBart and SPRING, both of which we reported results for. Cost for one run of LLM-R via GPT4o was \textasciitilde{}\$20, exceeding the non-existent cost of running only AMRBart on consumer PCs.

Although \llmr{} is designed to be model-agnostic, the prompt for Stage 2 LLM post-editor does not transfer reliably across parsers; thus the prompt must be tailored to the LLM for optimal performance. We defer studies on these limitation to current and ongoing research for parser domain generalizability and prompt interoperability.

\section*{Acknowledgments}

The icons in Figures~\ref{fig:sem} and~\ref{fig:sem-count} are the
\href{https://openmoji.org/library/emoji-1F916/}{robot}
by Mariella Steeb (OpenMoji), licensed under
\href{https://creativecommons.org/licenses/by-sa/4.0/}{CC BY-SA 4.0},
and the
\href{https://github.com/mozilla/fxemoji/blob/gh-pages/svgs/objects/u2702-blackscissors.svg}{scissors}
by Mozilla, licensed under
\href{https://creativecommons.org/licenses/by/4.0/}{CC BY 4.0}.

\bibliography{anthology,custom}

\appendix
\renewcommand{\thesection}{Appendix \Alph{section}}
\needspace{22\baselineskip}
\section*{Appendices}

\section{Dataset Breakdown}
\label{sec:datasets}
\begin{center}
\scriptsize
\setlength{\tabcolsep}{2pt}
\begin{tabular}{p{0.34\columnwidth}cp{0.40\columnwidth}}
    \hline
    \textbf{Subcorpus} & \textbf{N} & \textbf{Description} \\
    \hline
    long\_lists & 207 & Long coordination/enumerations. \\
    winograd & 150 & Winograd coreference pairs. \\
    deep\_recurs\_pronouns & 116 & Deep recursion + pronouns. \\
    deep\_recurs\_basic & 106 & Deep recursion structures. \\
    berts\_mouth & 95 & ``BERT's mouth'' constructions. \\
    deep\_recurs\_3s & 90 & Increased recursion depth. \\
    deep\_recurs\_rc\_coref & 75 & RC + contrastive coref. \\
    read\_by & 75 & Voice/arg-structure tests. \\
    give\_up\_in & 75 & Multiword predicates. \\
    see\_with & 75 & PP attachment/instrumental. \\
    deep\_recurs\_rc & 64 & Relative-clause embedding. \\
    nested\_control & 63 & Nested control structures. \\
    adjectives & 51 & Adjective modification. \\
    keep\_from & 50 & Control/PP-complement. \\
    bought\_for & 50 & PP/benefactive attachment. \\
    centre\_embedding & 43 & Center-embedded constructions. \\
    unseen\_senses & 41 & Unseen predicate senses. \\
    pp\_attachment & 30 & PP-attachment ambiguity. \\
    unseen\_roles & 17 & Unseen semantic roles. \\
    \hline
\end{tabular}
\captionof{table}{Synthetic subcorpora categories in GrAPES which target semantically complex phenomena taken from \citet{groschwitz-etal-2023-amr}.}
\label{tab:appendix-synthetic-subcorpora}
\end{center}

\begin{center}
\scriptsize
\setlength{\tabcolsep}{2pt}
\begin{tabular}{p{0.20\columnwidth}p{0.54\columnwidth}c}
    \hline
    \textbf{Dataset} & \textbf{Type} & \textbf{N} \\
    \hline
    BOLT DF MT & Online discussion forum text & 133 \\
    Weblog / WSJ & Newspaper, Wall Street Journal & 100 \\
    BOLT DF English & Online discussion forum text & 229 \\
    LORELEI & Low-resource language, news, social media, discussion forum, web crawl, code-switching & 527 \\
    Proxy reports & Narrative text from newswire, event-heavy, structured news-style & 823 \\
    Xinhua MT & Newspaper, translated Xinhua news & 86 \\
    \hline
\end{tabular}
\captionof{table}{AMR3 source datasets with text type and test counts \citep{Knight2020AMR3}.}
\label{tab:appendix-source-datasets}
\end{center}

\begin{center}
\footnotesize
\begin{tabular}{p{0.22\columnwidth}cp{0.42\columnwidth}}
    \hline
    \textbf{Category} & \textbf{Count} & \textbf{Example} \\
    \hline
    date\_only & 94 & 09/02/2010 20:51 \\
    date\_country & 69 & NEW DELHI, Aug. 22 \\
    category\_tags & 41 & proliferation; technology; international; politics \\
    id\_or\_code & 32 & SetC-SMS-093 \\
    country\_label & 19 & South Africa (ZA) \\
    formatting\ & 8 & ( End ) \\
    \hline
\end{tabular}
\captionof{table}{Non-sentence or metadata-like items observed in AMR3 \citep{Knight2020AMR3}.}
\label{tab:appendix-amr3-nondata}
\end{center}

\needspace{16\baselineskip}
\section{Linguistic Categories}
\begin{center}
\scriptsize
\setlength{\tabcolsep}{2pt}
\begin{tabular}{>{\raggedright\arraybackslash}p{0.23\columnwidth}>{\raggedright\arraybackslash}p{0.45\columnwidth}>{\raggedright\arraybackslash}p{0.25\columnwidth}}
    \hline
    \textbf{Category} & \textbf{Explanation (prompt Step~2.2)} & \textbf{Defect class (Step~2.4)} \\
    \hline
    Proposition-type shift & Identity/\allowbreak class/\allowbreak property $\leftrightarrow$ eventive/\allowbreak locative/\allowbreak role; necessity $\leftrightarrow$ possibility/\allowbreak ability. & Predicate-class / proposition-type mismatch \\ \noalign{\vskip 3pt}
    Predicate/\allowbreak sense drift & Wrong sense or near-synonym, e.g., ``require'' $\rightarrow$ ``allow'', ``fear'' $\rightarrow$ ``worry''. & Predicate-class / proposition-type mismatch \\ \noalign{\vskip 3pt}
    Core ARG errors & ARG0/ARG1 swapped, wrong anchoring, required ARG missing, spurious ARG. & Core ARG errors \\ \noalign{\vskip 3pt}
    Negation/\allowbreak modality scope & What is denied/\allowbreak required changes; negation attached to the wrong relation. & Negation/\allowbreak modality scope \\ \noalign{\vskip 3pt}
    Quantifier/\allowbreak number errors & Determiners, numerals, measurements, comparatives/\allowbreak superlatives, degree. & --- \\ \noalign{\vskip 3pt}
    Entity/\allowbreak name/\allowbreak grounding & Wrong entity type, broken names, incorrect dates/\allowbreak times, wrong units. & Entity structure + linking \\ \noalign{\vskip 3pt}
    Attachment errors & PP/\allowbreak adjunct (cause, condition, purpose, manner, time, location) attaches to the wrong predicate or argument. & Attachment errors \\ \noalign{\vskip 3pt}
    Reentrancy/\allowbreak coreference & Missing/\allowbreak wrong coreference (esp.\ across coordination/\allowbreak control), merged/\allowbreak split referents. & Reentrancy/\allowbreak coreference failures \\ \noalign{\vskip 3pt}
    Coordination/\allowbreak ellipsis & Conjunction scope, shared arguments not preserved, gapping/\allowbreak ellipsis not reconstructed. & Coordination / shared material / ellipsis \\ \noalign{\vskip 3pt}
    Polarity/\allowbreak contrast & ``Not X but Y'', concessives, contrastive coordination loses the opposition. & Negation/\allowbreak modality scope \\
    \hline
\end{tabular}
\captionof{table}{GAP categories of the \llmr{} prompt (M2, Box~\ref{box:m2-diagnosis} in \ref{app:prompts}), with the AMR defect class each is traced to in Box~\ref{box:m2-cause}. Step~2.4's ``structural generalization breakdowns'' (deep nesting, long lists) surface as Core ARG, Attachment, or Coordination errors. Table~\ref{tab:refiner-categories} uses these categories.}
\label{tab:appendix-linguistic-categories}
\end{center}

\section{\llmr{} Repairs}
\label{app:repairs}
\begin{center}
\scriptsize
\setlength{\tabcolsep}{2pt}
\begin{tabular}{p{0.39\columnwidth}p{0.30\columnwidth}rr}
    \hline
    \textbf{Theme} & \textbf{Category} & \textbf{GrAPES} & \textbf{AMR3} \\
    \hline
    arg\_roles\_structure         & Core ARG errors          & 23.0\% & 16.8\% \\ \noalign{\vskip 3pt}
    coordination\_ellipsis        & Coordination/ellipsis    & 15.0\% &  5.3\% \\ \noalign{\vskip 3pt}
    numbers\_quantifiers\_measure & Quantifier/number errors & 13.8\% & 10.6\% \\ \noalign{\vskip 3pt}
    attachment\_scope             & Attachment errors        &  12.5\% &  6.2\% \\ \noalign{\vskip 3pt}
    predicate\_sense\_frame       & Predicate/sense drift; Proposition-type shift & 8.5\% & 17.7\% \\ \noalign{\vskip 3pt}
    temporal\_tense\_time         & Entity/name/grounding (dates, times) &  6.5\% & 14.2\% \\ \noalign{\vskip 3pt}
    coreference\_pronoun          & Reentrancy/coreference   & 4.8\%  &  0.9\% \\ \noalign{\vskip 3pt}
    negation\_polarity            & Negation/modality scope; Polarity/contrast & 4.3\% & 3.5\% \\ \noalign{\vskip 3pt}
    named\_entity\_participant    & Entity/name/grounding (names, types, wiki) & 2.9\%  & 23.9\% \\ \noalign{\vskip 3pt}
    other / misc              & (none of Table~\ref{tab:appendix-linguistic-categories}) &  8.7\% &  0.9\% \\
    \hline
    Total Count           &                          & 456 & 113 \\
    \hline
\end{tabular}
\captionof{table}{Diagnosis categories (Table~\ref{tab:appendix-linguistic-categories}) of \textsc{LLM-R+L} repairs for one run of each dataset. Each sample is counted once, under the category of its first repair. \\
Repaired fraction: 456 (repaired) / 1471 (total GrAPES) $=$ 31\%; 113 (repaired) / 1898 (AMR3 sentences) $=$ 6.0\%. }
\label{tab:refiner-categories}
\end{center}

\needspace{12\baselineskip}
\section{Supplemental \llmrs{} Results}
\label{app:supplemental}

\begin{table}[!htbp]
\centering
\small
\setlength{\tabcolsep}{2.5pt}
\begin{tabular}{lccccc}
\hline
\textbf{Dataset} & \textbf{AMRBart} & \textbf{LLM-R} & $\Delta$\% & \textbf{LLM-R+L} & $\Delta$\% \\
\hline
GrAPES      & 0.80 & 0.82  & +2.1\% & 0.83  & +3.5\% \\
GrAPES$^R$  & 0.78 & 0.85  & +8.6\% & 0.87  & +12\%  \\
AMR3        & 0.85 & 0.852 & +0.2\% & 0.854 & +0.5\% \\
AMR3$^R$    & 0.81 & 0.82  & +0.8\% & 0.83  & +2.0\% \\
\hline
\end{tabular}
\caption{SMATCH++ \citep{opitz-2023-smatch} results comparing AMRBart, \llmrs{}, and \textsc{LLM-R+L}. $\Delta$\% is relative to AMRBart.}
\label{tab:smatchpp}
\end{table}

\begin{table}[!htbp]
\centering
\small
\setlength{\tabcolsep}{4.5pt}
\begin{tabular}{lccc}
\hline
\textbf{Metric} & \textbf{AMRBart} & \textbf{LLM-R} & \textbf{$\Delta$\%} \\
\hline
\noalign{\vskip 2pt}
Attribute         & 0.760 & 0.803 & +5.7\% \\
Unlabeled Edge    & 0.699 & 0.707 & +1.1\% \\
Unlabeled         & 0.831 & 0.840 & +1.1\% \\
Node (no senses)  & 0.882 & 0.890 & +1.0\% \\
No WSD            & 0.801 & 0.808 & +0.9\% \\
\hline
\end{tabular}
\caption{Fine-grained SMATCH sub-metrics (F1) \citep{damonte-etal-2017-incremental} for AMRBart and \llmr{}, with $\Delta$\% relative to AMRBart. Shown for one run of \llmr{} on the repaired subset of AMR3 (AMR3$^R$).}
\label{tab:fine-grained}
\end{table}

\begin{table*}[!htbp]
\centering
\small
\setlength{\tabcolsep}{4pt}
\resizebox{\textwidth}{!}{%
\begin{tabular}{lcccc|cccc}
\hline
\textbf{Dataset} &
\multicolumn{4}{c|}{\textbf{SemBLEU}} &
\multicolumn{4}{c}{\textbf{SMATCH}} \\
\cline{2-5} \cline{6-9}
& LLM-R $\Delta$\% & LLM-R CI & LLM-R+L $\Delta$\% & LLM-RL CI
& LLM-R $\Delta$\% & LLM-R CI & LLM-R+L $\Delta$\% & LLM-RL CI \\
\hline
GrAPES          
& +3.7\% & [3.3\%, 4.2\%] & +5.2\% & [4.3\%, 6.1\%] 
& +1.9\% & [1.5\%, 2.3\%] & +2.7\% & [2.3\%, 3.1\%] \\

GrAPES$^R$ 
& +16\% & [13\%, 18\%] & +22\% & [19\%, 24\%] 
& +8.4\% & [6.8\%, 9.0\%] & +11\% & [9.8\%, 12\%] \\

AMR3            
& +0.5\% & [0.3\%, 0.7\%] & +0.9\% & [0.8\%, 1.2\%] 
& +0.2\% & [0.1\%, 0.3\%] & +0.4\% & [0.2\%, 0.5\%] \\

AMR3$^R$   
& +2.9\% & [2.1\%, 4.2\%] & +5.1\% & [4.6\%, 5.5\%] 
& +0.7\% & [0.2\%, 1.4\%] & +1.9\% & [1.3\%, 2.2\%] \\
\hline
\end{tabular}%
}
\caption{AMRBart-relative improvements and confidence intervals for \llmrs{} and \textsc{LLM-R+L} across SemBLEU and SMATCH.}
\label{tab:metric-results-ci}
\end{table*}

\FloatBarrier

\section{Linguistic Analyzers}
\label{app:linguistic-context}

To support structured post-processing and error diagnosis, we augment parser outputs with auxiliary linguistic annotations produced by established NLP toolkits. These features provide complementary syntactic, semantic, and discourse-level signals that are not always explicitly encoded in AMR graphs but are useful for detecting inconsistencies and guiding repairs.

\paragraph{Semantic Role Labeling.}
We obtain predicate–argument structure using AllenNLP's semantic role labeling model. These annotations supply explicit argument spans and role labels, enabling cross-validation between predicted AMR argument structure and surface predicate realizations.

\paragraph{Constituency Structure.}
Phrase-level syntactic structure is extracted using spaCy in combination with the Benepar constituency parser. Constituency parses provide hierarchical phrase boundaries, which assist in identifying attachment ambiguities, coordination scope, and clause segmentation.

\paragraph{Part-of-Speech Tags.}
Lexical category information is obtained from spaCy POS tagging. These tags help detect lexical mismatches, such as when a predicted AMR concept corresponds to a different grammatical category than realized in the sentence.

\paragraph{Dependency Parses.}
We use spaCy dependency parses to capture grammatical relations between tokens. Dependency edges provide complementary structural information to AMR relations and are particularly useful for identifying subject–object inversions, modifier scope errors, and incorrect argument attachment.

\paragraph{Named Entities.}
Named entity recognition from spaCy is used to ground referential expressions and ensure that entity nodes in the AMR graph correspond to spans identified as proper names, locations, organizations, or other entity types.

\paragraph{Discourse Relations.}
To model inter-clausal semantics, we extract discourse relations using DiscoPy. These annotations identify rhetorical relations such as contrast, cause, and temporal ordering, which can inform higher-level graph structure and help diagnose missing or mis-specified discourse-level relations.

\paragraph{Role in the Pipeline.}
All auxiliary annotations are treated as deterministic external signals and are never modified during post-processing. Instead, they function as diagnostic evidence that informs repair decisions. This design preserves reproducibility while enabling structured comparison between predicted graphs and surface linguistic structure.

\begin{figure}[t]
    \centering
    \includegraphics[width=\columnwidth]{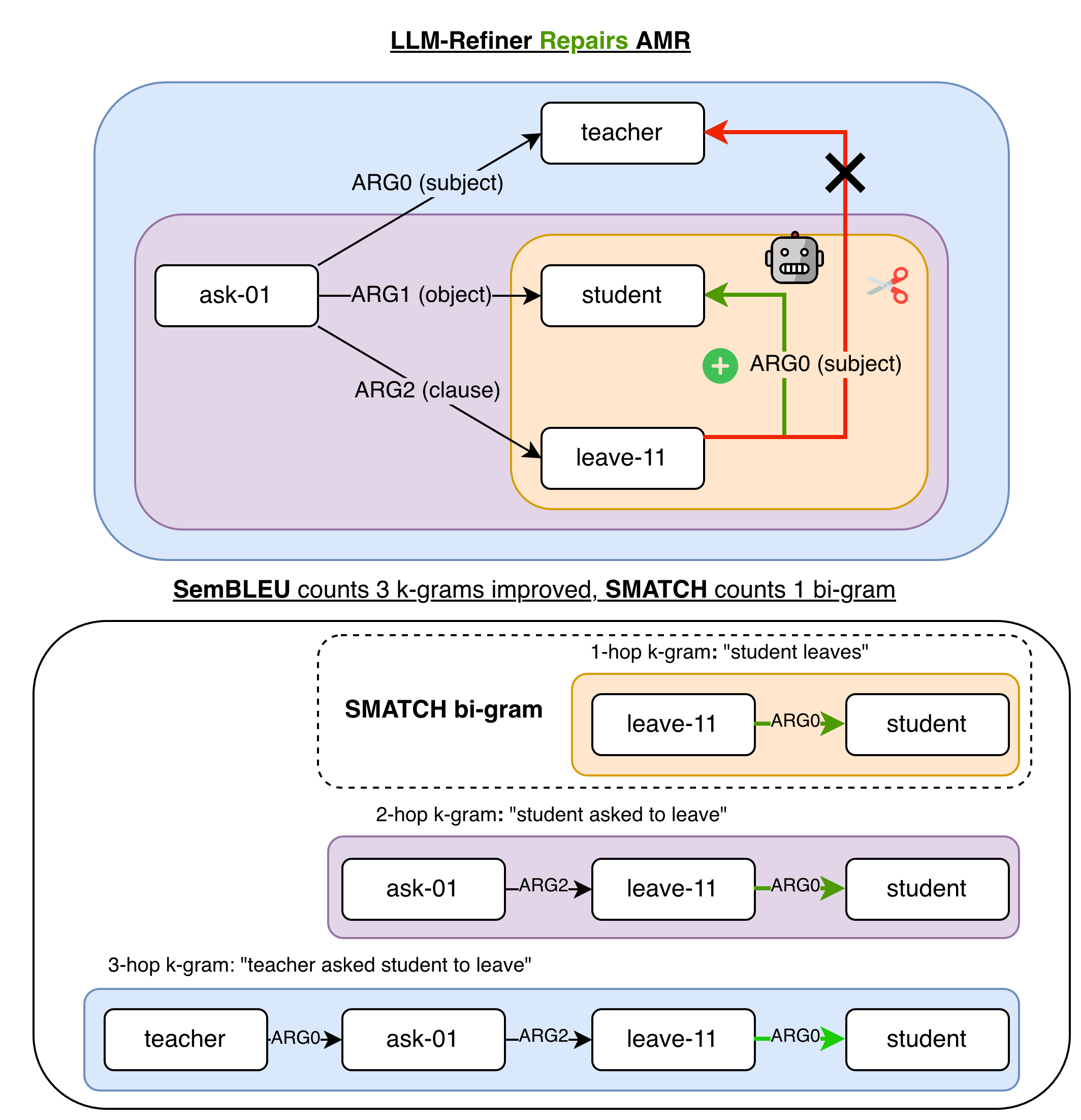}
    \caption{SMATCH considers improvement only from a single-hop perspective,
    ignoring the larger $k$-hop semantic context captured by SemBLEU
    (see also Fig.~\ref{fig:sem}).}
    \label{fig:sem-count}
\end{figure}

\section{\textit{Semantic Neighborhood} and \textit{Semantic Keystone} Case Study}
\label{sec:keystonecase}

For a given AMR element (node / edge), SemBLEU counts a larger radius of semantic impact than SMATCH does. Consider the example in Figs.~\ref{fig:sem} and \ref{fig:sem-count}: ``The teacher asked the student to leave.'' The AMRBart-produced graph mis-represents the subject of the ``leaving'' event, assigning it to the teacher instead of the student: \llmr{} removes the \textcolor{myred}{incorrect} edge and adds the \textcolor{mygreen}{correct} edge. While the repair only modifies a single edge, this repair simultaneously fixes the larger context: the \textcolor{mygold}{1-hop meaning ``student leaves''}, the \textcolor{mypurple}{2-hop meaning ``asked to leave''}, and the \textcolor{myblue}{3-hop meaning ``the teacher asked the student to leave''}. It is therefore a local graphical edit with wide semantic impact. 
We call this set of semantic meanings that an element affects its \semneighborhood{}s and said element a \semkeystone{} because of its semantic significance.

\section{\llmr{} Output Schema}
\label{app:json-schema}

Stage~2 returns one JSON object. Every stored output in our runs has exactly these fields; a sample has 0, 1, or 2 repairs.

\begin{tcolorbox}[title={Stage 2 JSON output (annotated)}]
\scriptsize
\begin{verbatim}
{ "output_amr": "<final AMR | original AMR>",
  "repairs": [            // 0-2 items; [] = no change
    { "target":         "<AMR node(s)/edge>",   // M2
      "diagnosis":      "<meaning lost/wrong>", // M2
      "applied_change": "<AMR edit>",  // M3
      "filtered_change": "number of edits discarded"}]} // M3
\end{verbatim}
\end{tcolorbox}

\texttt{target} and \texttt{diagnosis} record the M2 gap and its AMR locus; \texttt{applied\_change} records the M3 edit that is reflected in \texttt{output\_amr}. The M3 filter (gate, confidence $\geq 0.70$, at most two fixes) leaves no field of its own: candidates it discards are simply absent, and \texttt{repairs:\ []} means no candidate survived. The Validator then parses \texttt{output\_amr} with Penman (Sec.~\ref{sec:pipeline}).

\smallskip
\noindent\textit{Example (AMR3, one run).} Sentence: ``\ldots China is deeply concerned with the current situation in Iraq \ldots''
{\scriptsize
\begin{verbatim}
{ "target": "z4 / concern-01 :ARG0 and :ARG1",
  "diagnosis": "The roles of ARG0 and ARG1 for
    'concern-01' were swapped. China is the one
    concerned about the situation in Iraq.",
  "applied_change": "Swapped :ARG0 and :ARG1
    for 'concern-01'.",
  "filtered_change": 0  }
\end{verbatim}
}

\section{LLM-Refiner Prompt Excerpts}
\label{app:prompts}

Stage~2 is a single chat call per sample (GPT-4o, temperature~0, at most 2{,}000 output tokens, JSON response mode). The system prompt is Box~\ref{box:p-system}. The user prompt is Box~\ref{box:p-inputs} followed by  Boxes~\ref{box:step0}--\ref{box:p-output}. \textsc{LLM-R+L} and \textsc{LLM-R} share Boxes~\ref{box:step0}--\ref{box:p-output} and the output format word for word. They differ in two places only:
\begin{pitems}
  \item \textsc{LLM-R+L} adds the \lingcontextP{} inputs (Box~\ref{box:p-inputs}, item~4).
  \item \textsc{LLM-R} omits them and adds the evidence and prohibition rules of Box~\ref{box:p-system-r} to its system prompt.
\end{pitems}
Each prompt step from (Boxes~\ref{box:step0}--\ref{box:p-output}) is marked by its module number and titled to match the text in boxes for Fig.~\ref{fig:module}.

\begin{llmrpbox}{System prompt (\textsc{LLM-R+L})}
\label{box:p-system}
You are an expert Abstract Meaning Representation (AMR) parsing and evaluation specialist. Your task is to determine whether the CRITICAL meaning in the RAW SENTENCE is misrepresented in the GENERATED TEXT, to trace it to gaps in the INITIAL AMR, using linguistic context as supporting evidence, and to apply at most TWO high-confidence repairs if strictly necessary.

\smallskip
\textbf{Important assumptions:}
\begin{pitems}
  \item The input AMR is already high quality ($\approx$80\%+ F1 on average); most phenomena are already correctly represented. Your default action is NO CHANGE.
  \item RAW SENTENCE is the ground truth for truth-conditional meaning. GENERATED TEXT is primary evidence; use it to locate likely AMR defects.
  \item ``Critical meaning'' is meaning that would change the truth conditions, participant roles, scope, reference, or temporal grounding of the sentence.
\end{pitems}
\textbf{Only fix the INITIAL AMR when ALL are true:}
(1)~a specific meaning of the RAW SENTENCE is missing, distorted, or mis-scoped in the GENERATED TEXT and traced to a specific AMR node or structure;
(2)~it affects core truth-conditional meaning, event structure, reference, scope, or grounding, not stylistic realization or paraphrase;
(3)~it can be clearly explained with strong justifications;
(4)~the sentence unambiguously supports the missing meaning;
(5)~the repair does not remove existing correct meaning;
(6)~you are confident the AMR is wrong, not merely underspecified;
(7)~you can point to concrete lexical, structural, or referential evidence of a mismatch.

\smallskip
\textbf{Confidence threshold:} report an issue and repair only when confidence that the AMR is wrong is $\geq 0.70$; otherwise return the original AMR with \texttt{repairs:\ []} (format in Box~\ref{box:p-output}).
\end{llmrpbox}

\begin{llmrpbox}{System prompt additions (\textsc{LLM-R} only)}
\label{box:p-system-r}
The \textsc{LLM-R} system prompt is Box~\ref{box:p-system} without ``using linguistic context as supporting evidence'', plus:

\smallskip
\textbf{What counts as evidence.} Use the mismatch between the RAW SENTENCE (ground truth) and the GENERATED TEXT (diagnostic signal) as primary evidence, and keep it only if it traces to a specific AMR defect. A valid issue requires both:
(1)~the GENERATED TEXT fails to preserve a truth-conditional meaning of the RAW SENTENCE; and
(2)~the AMR has a graph explanation (missing, wrong, or incomplete node, edge, operator, or scope).
Graph-only evidence is allowed (e.g., a missing ARG role or required reentrancy). The default causal story is: \emph{sentence expresses X $\rightarrow$ text misses X $\rightarrow$ an AMR defect explains why.}

\smallskip
\textbf{Scope/operator check.} For negation, modality, quantification, or attribution, verify that the operator attaches to the correct semantic target (the denied, required, or scoped relation), not to an argument concept, unless the sentence expresses a negated concept (``non-X'', ``fearless''). A wrong target is a concrete AMR defect.

\smallskip
\textbf{You must NOT:}
\begin{pitems}
  \item repair stylistic, paraphrastic, or optional meaning;
  \item repair based solely on surface triggers or constructions, or when there is reasonable doubt;
  \item make speculative sense edits (except replacing a predicate of the wrong semantic class when clearly licensed);
  \item apply more than TWO repairs, invent meaning, or perform global rewrites.
\end{pitems}
Your goal is conservative correctness, not recall. If unsure, do nothing.
\end{llmrpbox}

\begin{llmrpbox}{User prompt: inputs}
\label{box:p-inputs}
\textbf{1 RAW SENTENCE:} \texttt{\{sentence\}}\\
\textbf{2 GENERATED TEXT:} \texttt{\{generated\_text\}} (Stage~1 AMR-to-text output)\\
\textbf{3 INITIAL AMR:} \texttt{\{amr\}} (Stage~1 parser output, PENMAN)\\
\textbf{4 LINGUISTIC CONTEXT} (\\
\hspace*{1em}Semantic Role Labeling \texttt{\{context.srl\}}; Constituency Parse \texttt{\{context.constituency\}}; POS Parse \texttt{\{context.pos\}}; Dependency Parse \texttt{\{context.dependency\}}; Discourse Relations \texttt{\{context.discourse\}}; Named Entity Recognition \texttt{\{context.ner\}}.
\end{llmrpbox}

\begin{llmrpbox}{M1 --- "Paraphrase"}
\label{box:step0}
\begin{pitems}
  \item Silently paraphrase the RAW SENTENCE, preserving truth conditions.
  \item Extract a compact semantic inventory:
  \begin{pitems}
    \item entities/participants, including abstract concepts, names, numbers, dates;
    \item predicates/relations: verbs, predicative adjectives, copulas, key prepositions and relational nouns;
    \item operators: negation, modality, quantifiers/degree, coordination.
  \end{pitems}
  \item Organize the meaning into truth-conditional bullets, one proposition each, noting:
  \begin{pitems}
    \item proposition type (identity/class/property vs.\ eventive vs.\ modal/attitude);
    \item argument orientation (avoid role reversal);
    \item attachment (what modifies what).
  \end{pitems}
  \item Mark any required coreference or sharing across bullets.
\end{pitems}
You will not output these bullets --- they guide later decisions.
\end{llmrpbox}

\begin{llmrpbox}{M1 --- "Create Semantic Intake"}
\label{box:step1-reconstruction}
Silently reconstruct the full intended meaning of the RAW SENTENCE into bullet points, covering:
\begin{pitems}
  \item \textbf{Predicates / events / states} and their intended senses.
  \item \textbf{Copular and classificational structures}, labeled as identity/equivalence (`X is (not) Y'), class membership (`X is a Y'), property attribution (`X is Adj'), or eventive relation (`X does Y').
  \item \textbf{Core roles (ARG0--ARGn)}: who did what to whom, role orientation (no role reversal), voice/alternations, implicit arguments.
  \item \textbf{Participants/entities}: typing, names, grounding (dates/times, numeric entities).
  \item \textbf{Modifiers and attachment}: adjectival vs.\ predicative structure, PP attachment, clause-level adjuncts (cause/purpose/condition).
  \item \textbf{Coordination, sharing, coreference}: required reentrancy/anaphora (entity or event).
  \item \textbf{Polarity, modality, illocutionary force, factuality}: scope, questions, attitude markers.
  \item \textbf{Temporal meaning}: tense/aspect, temporal relations (before/after/while), duration/frequency.
  \item \textbf{Quantification and degree}: quantifier type, numerals/measurement, comparatives/superlatives.
  \item \textbf{Presuppositions, implicatures, attribution}: stance, evidentials, speaker commitment, only if explicitly expressed.
\end{pitems}
You will not display these bullets --- you only use them.
\end{llmrpbox}

\begin{llmrpbox}{M2 --- "Compare Semantic Intake"}
\label{box:m2-diagnosis}
Combine the bullets of Steps~0--1 and compare them one by one against the GENERATED TEXT, using the AMR as the causal explanation for any divergence. For each bullet:

\smallskip
\textbf{2.1} Locate the span(s) of the GENERATED TEXT that should express it.

\smallskip
\textbf{2.2} Mark \textbf{PRESENT} only if the text keeps the same truth-conditional proposition:
\begin{pitems}
  \item the same proposition type;
  \item the same argument structure and orientation;
  \item the same operator scope (negation, modality, quantifiers, coordination);
  \item the same reference structure (required coreference/reentrancy);
  \item the same grounded content (entities, names, numbers, dates, degree).
\end{pitems}
Mark \textbf{GAP} if any critical meaning is missing, altered, or incompletely realized. Common GAP patterns (not exhaustive):
\begin{pitems}
  \item \textbf{Proposition-type shift:} identity/class/property rendered as eventive/locative/role (or vice versa); necessity $\leftrightarrow$ possibility/ability.
  \item \textbf{Predicate/sense drift:} wrong sense or near-synonym (``require'' $\rightarrow$ ``allow'', ``fear'' $\rightarrow$ ``worry'').
  \item \textbf{Core ARG errors:} ARG0/ARG1 swapped, wrong anchoring, required ARG missing, spurious ARG.
  \item \textbf{Negation/modality scope:} what is denied/required changes; negation on the wrong relation.
  \item \textbf{Quantifier/number errors:} determiners, numerals, measurements, comparatives/superlatives, degree (``some'' $\rightarrow$ ``all'', ``at least'' $\rightarrow$ ``exactly'').
  \item \textbf{Entity/name/grounding:} wrong entity type, broken names, missing/incorrect dates or times, wrong units.
  \item \textbf{Attachment errors:} PP/adjunct (cause, condition, concession, purpose, manner, time, location) on the wrong predicate/argument.
  \item \textbf{Reentrancy/coreference:} missing/wrong coreference (esp.\ across coordination/control), merged/split referents.
  \item \textbf{Coordination/ellipsis:} conjunction scope, shared arguments not preserved, gapping/ellipsis not reconstructed.
  \item \textbf{Polarity/contrast:} ``not X but Y'', concessives, contrastive coordination loses the opposition.
\end{pitems}
Lexical overlap alone is never sufficient to mark PRESENT; truth conditions, scope, roles, and reference must match.
\end{llmrpbox}

\begin{llmrpbox}{M2 --- "Diagnose AMR"}
\label{box:m2-cause}
\textbf{2.3} For each GAP:
\begin{pitems}
  \item if the AMR licenses the GENERATED TEXT (predicate choice, role structure, operator class), it is an \textbf{AMR defect};
  \item if the text contradicts the AMR but matches the RAW meaning, it is a \textbf{generator error} --- discard;
  \item if ambiguous, default to acceptable --- discard.
\end{pitems}
\textbf{2.4} Keep the issue only if you can point to one cause, such as:
\begin{pitems}
  \item \textbf{Predicate-class / proposition-type mismatch} (eventive vs.\ copular; wrong semantic field);
  \item \textbf{Core ARG errors} (inversion, anchoring, missing core role, spurious ARG);
  \item \textbf{Negation/modality scope} (\texttt{:polarity -} or modal force on the wrong node);
  \item \textbf{Reentrancy/coreference failures} (control, coordination sharing, pragmatic coreference);
  \item \textbf{Entity structure + linking} (broken \texttt{:name}, entity type, date/number format, \texttt{:wiki});
  \item \textbf{Attachment errors} (esp.\ PP attachment);
  \item \textbf{Coordination / shared material / ellipsis};
  \item \textbf{Structural generalization breakdowns} (nested control/coordination, center embedding, CP recursion, long lists, stacked adjectives).
\end{pitems}
\end{llmrpbox}

\begin{llmrpbox}{M3 --- "Prune"}
\label{box:m3-repair}
List all candidate AMR issues, then immediately prune using a strict gate.

\smallskip
\textbf{Keep an issue only if all are true:}
\begin{pitems}
  \item \textbf{Critical:} it changes truth conditions, proposition type, argument structure, or scope.
  \item \textbf{AMR-visible:} the exact node(s) and edge(s) responsible can be identified.
  \item \textbf{Unambiguous:} the RAW SENTENCE clearly licenses only one correction.
  \item \textbf{High-confidence and regression-safe:} fixing it removes no other correct meaning.
\end{pitems}
\textbf{Discard immediately if any are true:}
\begin{pitems}
  \item stylistic, optional, or paraphrastic;
  \item depends on world knowledge or discourse inference;
  \item requires changing unrelated predicates, senses, or conjuncts;
  \item the sentence plausibly supports multiple analyses.
\end{pitems}
If no issues survive, return the unchanged AMR with \texttt{repairs:\ []}.
\end{llmrpbox}

\begin{llmrpbox}{M3 --- "Group + Rank + Eliminate"}
\label{box:m3-select}
From the remaining issues:
\begin{pitems}
  \item Select at most \textbf{two} highest-confidence fixes.
  \item A fix is eligible only if the AMR edit itself is high-confidence, not just the diagnosis: name the exact node(s)/edge(s) to change and the semantic relation class the edit must realize (identity/class/property vs.\ eventive vs.\ locative/role/temporal).
  \item If several symptoms arise from one defect, consolidate them into \textbf{one} fix.
  \item Reject a fix if several predicate/operator classes are plausible, or if it is a ``label swap'' without clear justification from the RAW SENTENCE.
  \item If confidence in the edit's semantic correctness is not extremely high, return the unchanged AMR with \texttt{repairs:\ []}.
\end{pitems}
\end{llmrpbox}

\begin{llmrpbox}{M3 --- "Apply Fix"}
\label{box:m3-apply}
Apply the targeted AMR edits that directly resolve the selected issues.

\smallskip
\textbf{Constraints:}
\begin{pitems}
  \item Introduce no meaning, and no change of proposition type, beyond what the RAW SENTENCE licenses.
  \item Do not restructure unrelated conjuncts or predicates.
  \item Preserve all participants, modifiers, and operator-scope relations not implicated by the fix.
\end{pitems}
\textbf{Mandatory internal validation after each edit:}
\begin{pitems}
  \item re-derive the truth-conditional content of the edited subgraph and confirm its proposition type;
  \item confirm negation/modality scope attaches to the intended relation;
  \item confirm previously satisfied bullets remain satisfied and no new relations are introduced.
\end{pitems}
Each \texttt{applied\_change} must correspond exactly to the edit in \texttt{output\_amr} and name the edited locus (node id + edge/predicate) and the relation class achieved.
\end{llmrpbox}

\begin{llmrpbox}{Output format (strict JSON)}
\label{box:p-output}
Return only one JSON object.

\smallskip
If at least one repair was applied:
{\scriptsize
\begin{verbatim}
{"output_amr": "<final AMR>",
 "repairs": [
   {"target": "<AMR node(s) or token span>",
    "diagnosis": "<what critical meaning
                  was missing/wrong>",
    "applied_change": "<AMR edit>",
    "filtered_change": <number of changes filtered>}]}
\end{verbatim}
}
If no issue was warranted:
{\scriptsize
\begin{verbatim}
{"output_amr": "<original AMR>",
 "repairs": []}
\end{verbatim}
}
\end{llmrpbox}

\begin{llmrpbox}{Validator: structural repair prompt}
\label{box:p-validator}
\textbf{Check.} \texttt{output\_amr} must decode and re-encode with Penman library. If both pass, it is accepted unchanged; otherwise:

\smallskip
\textbf{System:} You are an AMR structural repair specialist. You receive an AMR graph that may have only structural or formatting errors (e.g., unbalanced parentheses, invalid role labels, duplicated variable names) together with a short error message from a strict parser. Fix only these structural issues while preserving the graph's nodes, roles, and overall meaning as much as possible. Do not change the semantics beyond what is needed to make the graph well-formed. Respond with a strict JSON object containing a single field \texttt{fixed\_amr}.

\smallskip
\textbf{User (JSON):} \texttt{amr\_with\_issues}; \texttt{error\_hint} (one-line summary of the parser error); \texttt{instructions}: ``Return a structurally valid AMR as `fixed\_amr' that preserves the original graph as much as possible, fixing only structural and formatting problems indicated by the error hint.''

\smallskip
The check is repeated after each attempt, for up to 3 attempts (GPT-4o, temperature 0.8). If all fail, the sample is marked invalid.
\end{llmrpbox}

\end{document}